\def\year{2022}\relax
\documentclass[letterpaper]{article} 
\usepackage{aaai22}  
\usepackage{times}  
\usepackage{helvet}  
\usepackage{courier}  
\usepackage[hyphens]{url}  
\usepackage{graphicx} 
\usepackage{natbib}  
\usepackage{caption} 
\DeclareCaptionStyle{ruled}{labelfont=normalfont,labelsep=colon,strut=off} 
\usepackage{soul}
\usepackage{bbding} 
\usepackage{amssymb}
\usepackage{amsmath}
\usepackage{subfigure}
\usepackage{color}
\usepackage[switch]{lineno}  %
\newlength\savewidth\newcommand\shline{\noalign{\global\savewidth\arrayrulewidth
  \global\arrayrulewidth 1pt}\hline\noalign{\global\arrayrulewidth\savewidth}}
\title{Temporal Action Proposal Generation with Background Constraint}
\author{Haosen Yang$^{1,2}$\thanks{ Co-first authorship. This work was done when Haosen was a research intern at Baidu Inc, Wenhao was the project leader.},
        Wenhao Wu$^{2*}$,
        Lining Wang$^{1}$,
        Sheng Jin$^{1}$,\\
        Boyang Xia$^{2,3}$,
        Hongxun Yao$^{1}$\thanks{Corresponding author},
        Hujie Huang$^{1}$
        }
\affiliations{
   \textsuperscript{\rm 1} Harbin Institute of Technology  \\
    \textsuperscript{\rm 2} Department of Computer Vision Technology (VIS), Baidu Inc.\\
     \textsuperscript{\rm 3} University of Chinese Academy of Sciences
    \\
}

\begin{document}

\maketitle

\begin{abstract}
Temporal action proposal generation (TAPG) is a challenging task that aims to locate action instances in untrimmed videos with temporal boundaries.
To evaluate the confidence of proposals, the existing works typically predict action score of proposals that are supervised by the temporal Intersection-over-Union (tIoU) between proposal and the ground-truth.
In this paper, we innovatively propose a general auxiliary \emph{Background Constraint} idea to further suppress low-quality proposals, by utilizing the background prediction score to restrict the confidence of proposals. In this way, the Background Constraint concept can be easily plug-and-played into existing TAPG methods (\emph{e.g.}, BMN, GTAD). 
From this perspective, we propose the Background Constraint Network (BCNet) to further take advantage of the rich information of action and background. Specifically, we introduce an Action-Background Interaction module for reliable confidence evaluation, which models the inconsistency between action and background by attention mechanisms at the frame and clip levels.
Extensive experiments are conducted on two popular benchmarks, \emph{i.e.}, ActivityNet-1.3 and THUMOS14. The results demonstrate that our method outperforms state-of-the-art methods. Equipped with the existing action classifier, our method also achieves remarkable performance on the temporal action localization task.
\end{abstract}
\vspace{-0.3cm}
\begin{figure}[t]
    \centering
    \includegraphics[width=0.48\textwidth]{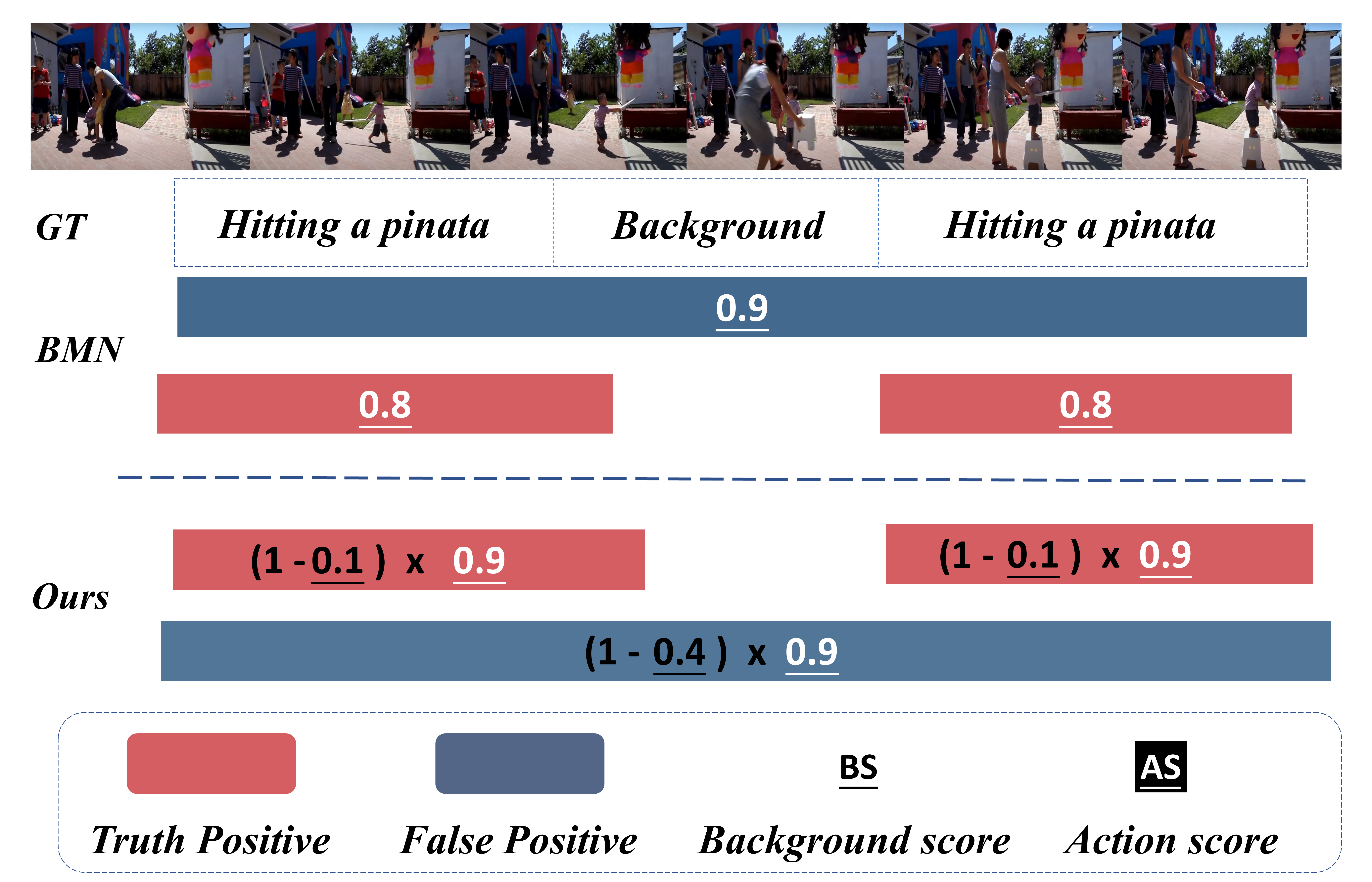}
    \vspace{-0.3cm}
    \caption{Illustration of the background constraint concept. We introduce a background score for the confidence of the proposal, which helps to restrict false-positive proposals.}
    \label{fig:teaser}
    \vspace{-0.3cm}
\end{figure}

\section{Introduction}
With the rapid development of mobile devices and the Internet, a massive amount of video content is being uploaded to the Internet every second. 
The volume of video information has far exceeded the processing capacity of the conventional manual system, thus video content analysis has attracted the extensive interest of academic and industrial communities.

One of the most active research topics in video understanding is temporal action detection, which focuses on both classifying the action instances present in an untrimmed video and localizing them with temporal boundaries.
The temporal action detection task, like object detection, is divided into two parts: temporal action proposal generation (TAPG) and action recognition.
Deep learning has recently been shown to significantly improve action recognition performance \cite{two-stream,tsm,wu2020MVFNet}.
However, the performance of the two-stage temporal action detectors in mainstream benchmarks~\cite{thumos, activitynet} still has much room for improvement, which is mostly influenced by the quality of proposals from temporal action proposal generation.

Hence, great efforts have been devoted to TAPG task~\cite{lin2018bsn, lin2019bmn}.
These research generally use the temporal Intersection-over-Union (tIoU) between the proposal and instance, called the action score, to evaluate the confidence of the proposal in order to develop high-quality temporal action proposals with dependable confidence scores.
However, the background information is also significant but was previously overlooked.
For instance, as illustrated in Figure~\ref{fig:teaser}, we can easily restrict false-positive proposals by detecting the background ``Move a chair".
Furthermore, we can evaluate the inconsistency between the background ``Move a chair" and the action ``Hitting a pinata", leading to better action score and background score of the proposal.
Motivated by above observations, 
we propose a general auxiliary \emph{Background Constraint} idea to reduce localization errors. Specifically, we introduce a background score for the proposal's confidence, and its supervision signal is defined by the temporal Intersection-over-Anchor (tIoA) between the proposal and the background. This concept can be flexibly integrated into existing TAPG methods (\emph{e.g.}, BMN~\cite{lin2019bmn}, GTAD~\cite{xu2020g}) to improve the performance in a plug-and-play fashion.

To further mine the rich information of action and background, in this paper, we propose the \emph{\textbf{B}ackground \textbf{C}onstraint \textbf{Net}work} (BCNet) to generate high-quality temporal action proposals. An essential component of BCNet is the \textit{Action-Background Interaction} (ABI) module, which performs both frame-level and clip-level action-background interaction to obtain reliable confidence scores of proposals. 
To do so, we first generate action features and background features for each frame using self-attention and difference-attention. Sliding windows are then used to generate multi-scale anchors from action and background features. The clip-level interaction then discovers the complex relationships between action-anchors and background-anchors, and outputs the action and background scores for these anchors.
We also propose a \emph{Boundary Prediction} (BP) module for precisely locating action boundaries. To capture the complex long-term temporal relationships while avoiding the influence of global noise, we aggregate the original feature sequence using self-attention and cross-attention mechanisms.
The output representation is then used as the global representation for the boundary prediction task.
Finally, we feed the boundary probabilities, action scores and background scores into the post processing module to get the final proposal set.

Experimental results show the superiority of our system on two popular datasets, \emph{i.e.,} ActivityNet \cite{caba2015activitynet} and THUMOS14 \cite{thumos}. Our BCNet achieves significant performance and outperforms existing state-of-the-art methods on both datasets. 
Our contributions are summarized as follows:
\begin{itemize}
    \item We introduce a Background Constraint concept, which can be integrated easily with existing TAPG methods (\emph{e.g.}, BMN, GTAD) and improve performance significantly.
    \item We propose a Background Constraint Network, which consists of multiple attention units, \emph{i.e.}, self-attention, cross-attention and difference-attention, and generates high-quality proposals by exploiting inconsistency between action and background.
    \item Extensive experiments demonstrate that our method outperforms the existing state-of-the-art methods on THUMOS14 and achieves comparable performance on ActivityNet-1.3, in both temporal action proposal generation task and temporal action detection task.
\end{itemize}

\section{Related Work}
\subsection{Video Action Recognition}
Action recognition is a fundamental task in the video understanding area.
Currently, there are two types of end-to-end action recognition methods: 3D CNN-based methods and 2D CNN-based methods.
3D CNNs are natural extensions of their 2D counterparts and are intuitive spatiotemporal networks that directly tackle 3D volumetric video data~\cite{c3d,i3d} but have a high computational cost. 
Other alternative efficient architectures, such as TSM~\cite{tsm}, TEI~\cite{teinet},  MVFNet~\cite{wu2020MVFNet}, DSANet~\cite{dsanet}, \emph{etc}, have been developed to capture temporal information with reasonable training resources.
These methods aim to design efficient temporal modules to perform efficient temporal modeling.
There is also ongoing research into dynamic inference~\cite{wu2020dynamic}, adaptive frame sampling techniques~\cite{wu2019multi,korbar2019scsampler}, which we believe can complement the end-to-end video recognition approaches.

\subsection{Temporal Action Proposal Generation}
Temporal action proposal generation aims to detect action instances with temporal boundaries and confidence in untrimmed videos.
Existing methods can be mainly divided into Top-down and Bottom-up methods. 
The Top-down methods~\cite{oneata2014lear,gao2017turn,gao2018ctap,gao2020accurate,chen2019relation} generate proposals using sliding windows or pre-defined anchors.
The Bottom-up methods mainly focus on evaluating ``actionness'', which indicates the probability of a potential action, for each frame or clip in a video.
These works~\cite{shou2016temporal, zhao2017temporal} use snippet-wise probability to generate candidate proposals.
BSN \cite{lin2018bsn} first proposes to predict start, end and actionness of each frame, then proposals are generated by constructing start and end points with high probabilities, with low confidence ones further abandoned by an evaluation module.
\cite{lin2019bmn,su2020bsn++,xu2020g,lin2020fast} generate all possible combinations of temporal locations to evaluate confidence of proposals.
\cite{liu2019multi} generates coarse segment proposals by perceiving the whole video sequence and predicts the frame actionness by densely evaluating each video frame.
These methods evaluate action scores of proposals with rich clip-level context.
However, these methods fail to take full advantage of background by focusing only on the action score.
In our work, we predict the extra background score for the confidence of proposals to reduce low-quality proposals.


\begin{figure*}[t]
    \centering
    \includegraphics[width=1\textwidth]{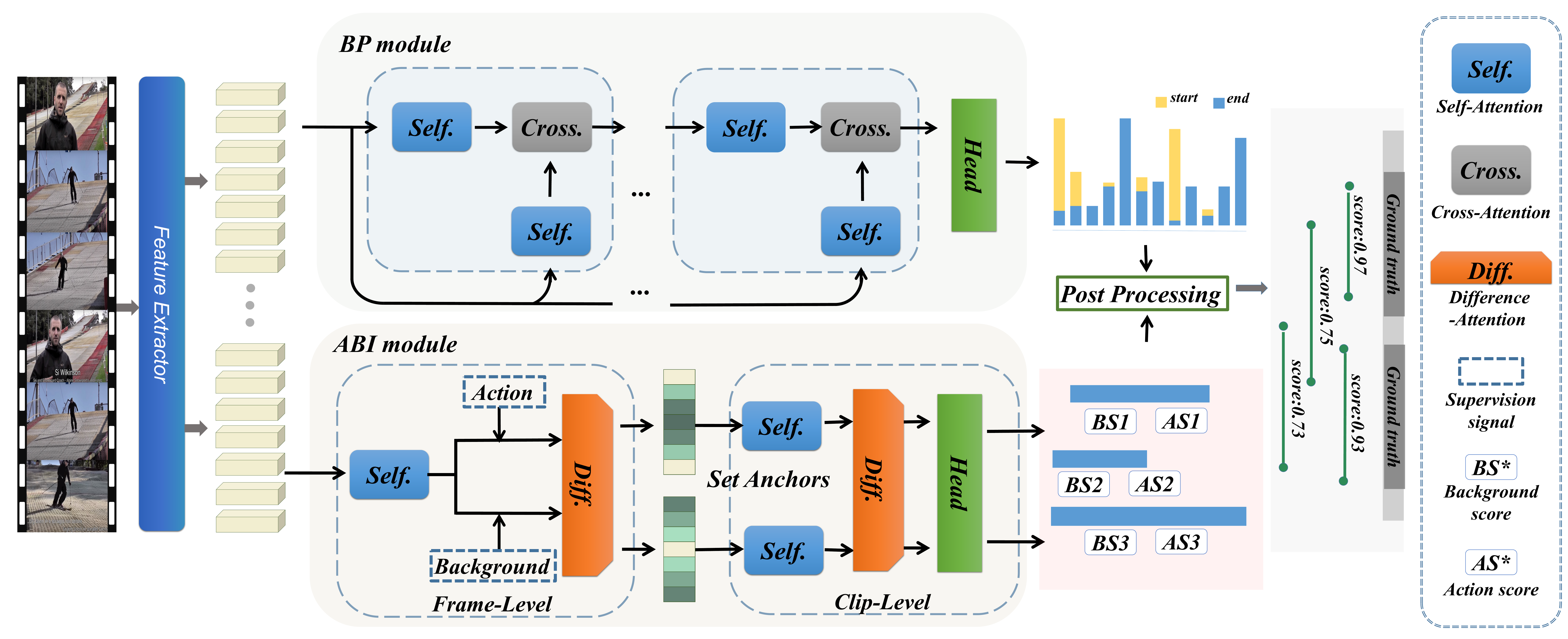}
      \caption{Illustration of the proposed Background Constraint Network. First we apply the feature extractor to encode video frames. Boundary Prediction (BP) module takes the feature sequence as input, and outputs boundary probability sequence. Action-Background Interaction (ABI) module takes the feature sequence as input, and outputs features of action and background at the frame level. Then, we set anchors on features of action and background, and feed them into clip-level interaction to generate action and background scores of anchors. Finally, we construct proposals based on boundary probabilities sequence and refine them using the corresponding anchor.}
      \vspace{-0.3cm}
    \label{fig:framework}
\end{figure*}

\subsection{Transformer and self-attention mechanism}
Transformers \cite{vaswani2017attention} has achieved great success in natural language processing.
Transformer architectures are based on a self-attention mechanism that summarizes content from the source sequence and is capable of modeling complex and arbitrary dependencies within a limited number of layers.
Recently, many works~\cite{dosovitskiy2020image,carion2020end,liu2021swin,tan2021relaxed,wang2021temporal} have revealed the great potential of Transformers in the computer vision task.
Inspired by the successful application of Transformers in various fields, we intuitively take advantage of Transformers in modeling long-range contextual information. In this paper, we utilize the Transformer-alike structure to devise three attention units.


\subsection{Background Modeling on Temporal Action Localization} 
Background modeling in Temporal action localization has received some attention.
Several previous works~\cite{shou2016temporal,yuan2016temporal} generate proposals by sliding window and classify them into $C+1$ classes for $C$ action classes plus background class. 
Also, several studies attempt to explicit background modeling for weakly-supervised temporal action localization.
Some works \cite{nguyen2019weakly,lee2020background} try to classify background frames as a separate class.
\cite{lee2020weakly} formulates background frames as out-of-distribution samples.
Essentially, all the above works aim to perform classification for these proposals.
Unlike them, in our work, we propose a Background Constraint concept to predict an additional background score for proposal confidence evaluation. To supervise the background score, we use temporal Intersection-over-Anchor (tIoA) between the proposal and the background. Our work concentrates on utilizing the background prediction score to restrict the confidence of proposals.

\section{Background Constraint Network}

As shown in Figure~\ref{fig:framework}, we propose a Background Constraint Network (BCNet) to generate high-quality proposals, which mainly consists of two main modules:  Action-Background Interaction Module and Boundary Prediction Module. Firstly, the Action-Background Interaction (ABI) module is adopted to perform both frame-level and clip-level action-background interaction to obtain reliable confidence scores of proposals. The Boundary Prediction (BP) module is then utilized to locate the boundaries of the proposals by exploiting complex long-term temporal relationships for boundary regression.

\subsection{Problem Definition}
\label{sec:definition}
An untrimmed video $U$ can be denoted as a frame sequence $U=\{u_t\}_{t=1}^{l_v}$ with $l_v$ frames, where $u_t$ denotes the $t$-th RGB frame of video $U$. 
The temporal annotation set of $U$ is made up of a set of temporal action instances as $\Psi_g=\{\varphi_n^g\}_{n=1}^{N_g}$ and $\varphi_n^g = (t_{s_n}, t_{e_n})$, where $N_g$ is the number of ground-truth action instances, $t_{s_n}$ and $t_{e_n}$ are the starting and ending time of the action instance $\phi_n^g$, respectively. 
During training phase, the $\Psi_g$ is provided.
While in the testing phase, the predicted proposal set $\Psi_p$ should cover the $\Psi_g$ with high recall and high temporal overlapping.

\subsection{Background Constraint}
To evaluate the confidence of the proposal, existing methods primarily use the temporal Intersection-over-Union (tIoU) between the proposal and instance, called \emph{action score}.
The temporal Intersection-over-Union (tIoU) is used to define the label of action score, which can be computed by:
\begin{equation}
    A_{label} = max \{ {\begin{vmatrix} \frac{G_i \cap P}{G_i \cup	P} \end{vmatrix}}_{i=1}^n\},
\end{equation}
where $G_i$ is the $i$-th ground truth and $P$ is the proposal, $n$ is number of ground truth.
In this paper, we propose a novel Background Constraint concept to suppress low-quality proposals (false positive proposal).
Specifically, we predict a extra \emph{background score} for evaluating the confidence of the proposal besides the action score.
The label of the background score is generated using temporal Intersection-over-Anchor (tIoA), which can be computed by:
\begin{equation}
    B_{label} = 1 - {\sum_{i=1}^{n} \frac{G_i \cap P}{P} },
\end{equation}
where $G_i$ is the $i$-th ground-truth and $P$ is the proposal, $n$ is number of ground-truth.
It is worth reminding that $A_{label}$ plus $B_{label}$ is not equal to 1 in most circumstances.

\begin{figure}[t]
    \centering
    \subfigure[Self-attention]{
    \includegraphics[width=0.135\textwidth]{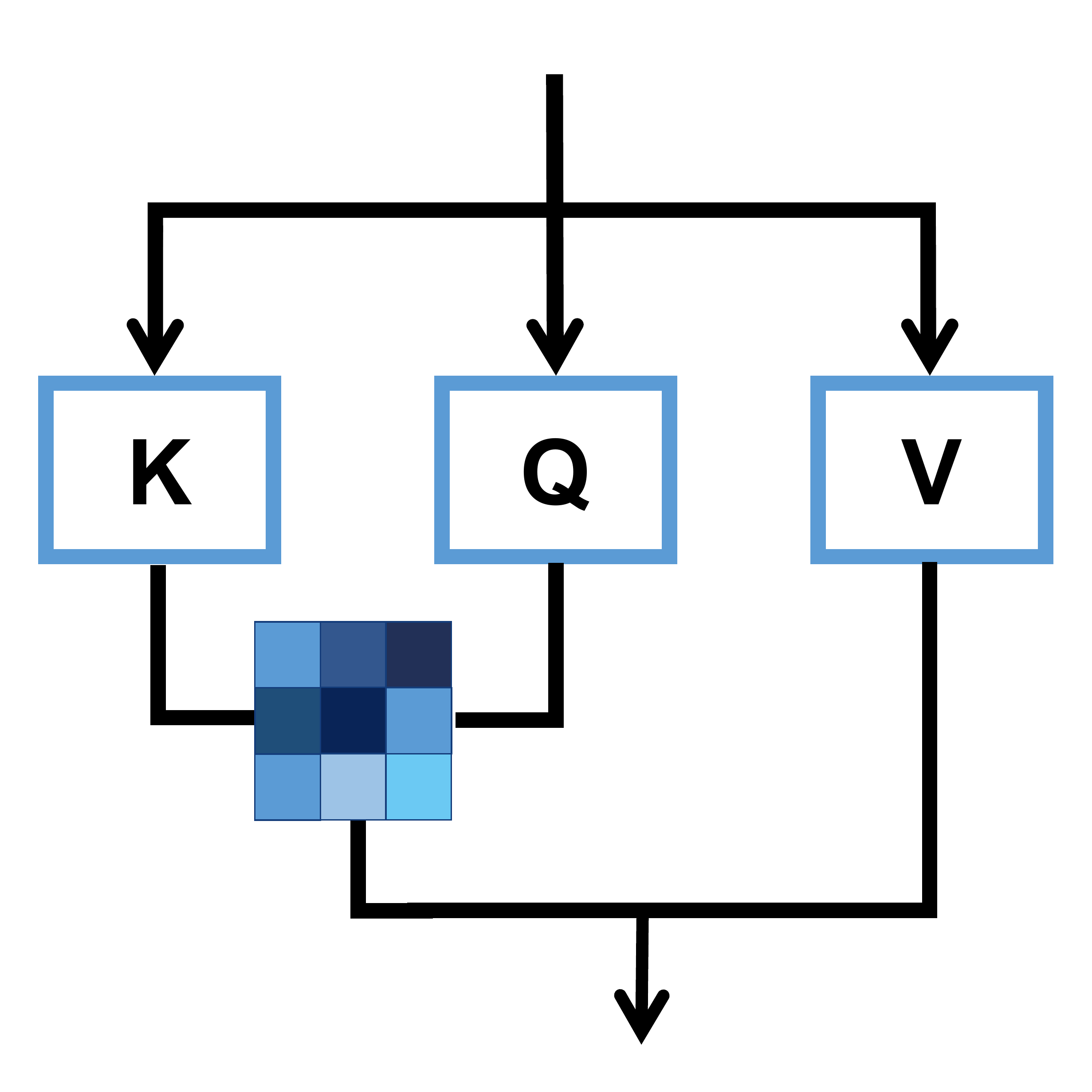}
    \label{fig:self}
    }
    \subfigure[Cross-attention]{
    \includegraphics[width=0.135\textwidth]{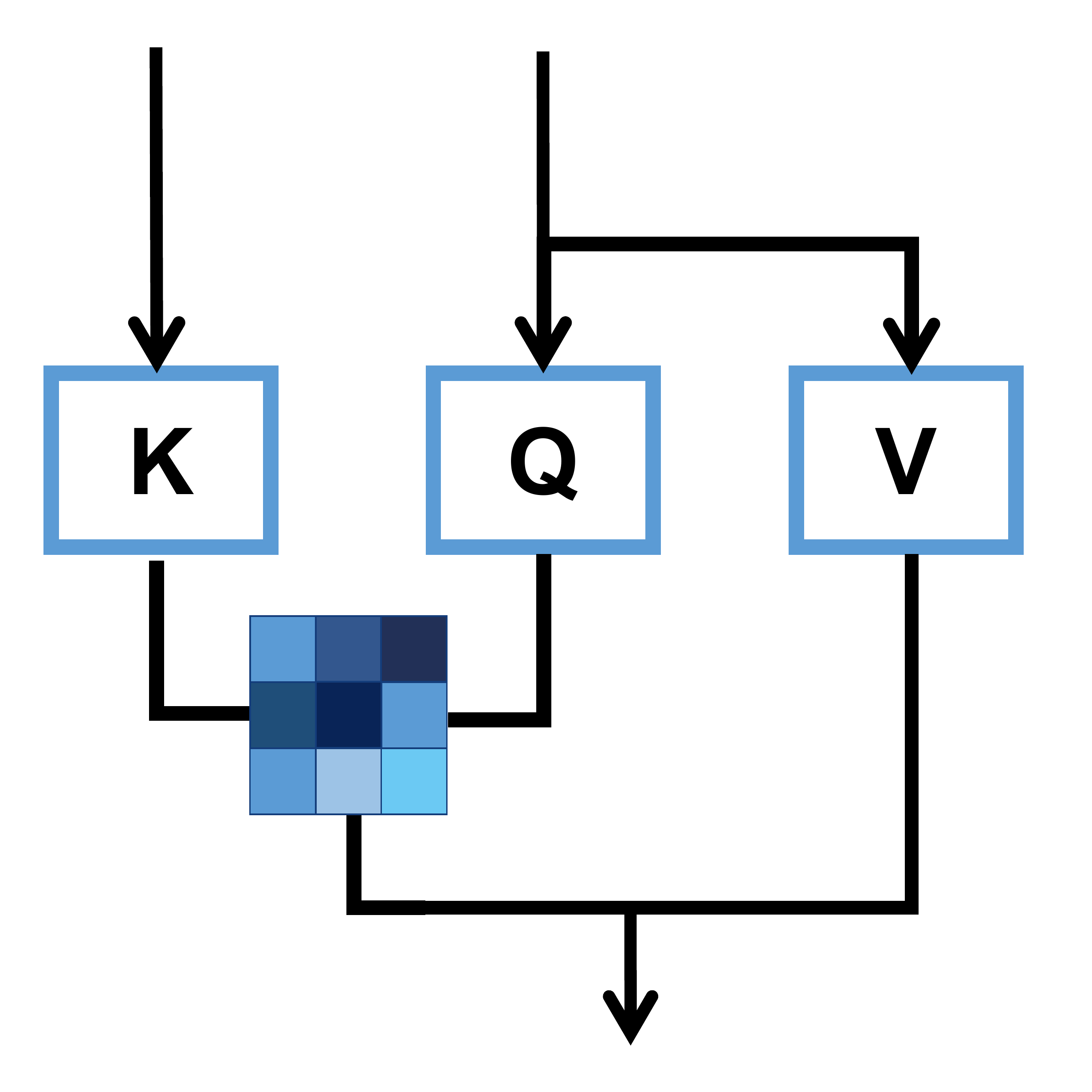}
    \label{fig:cross}
    }
    \subfigure[Difference-attention]{
    \includegraphics[width=0.162\textwidth]{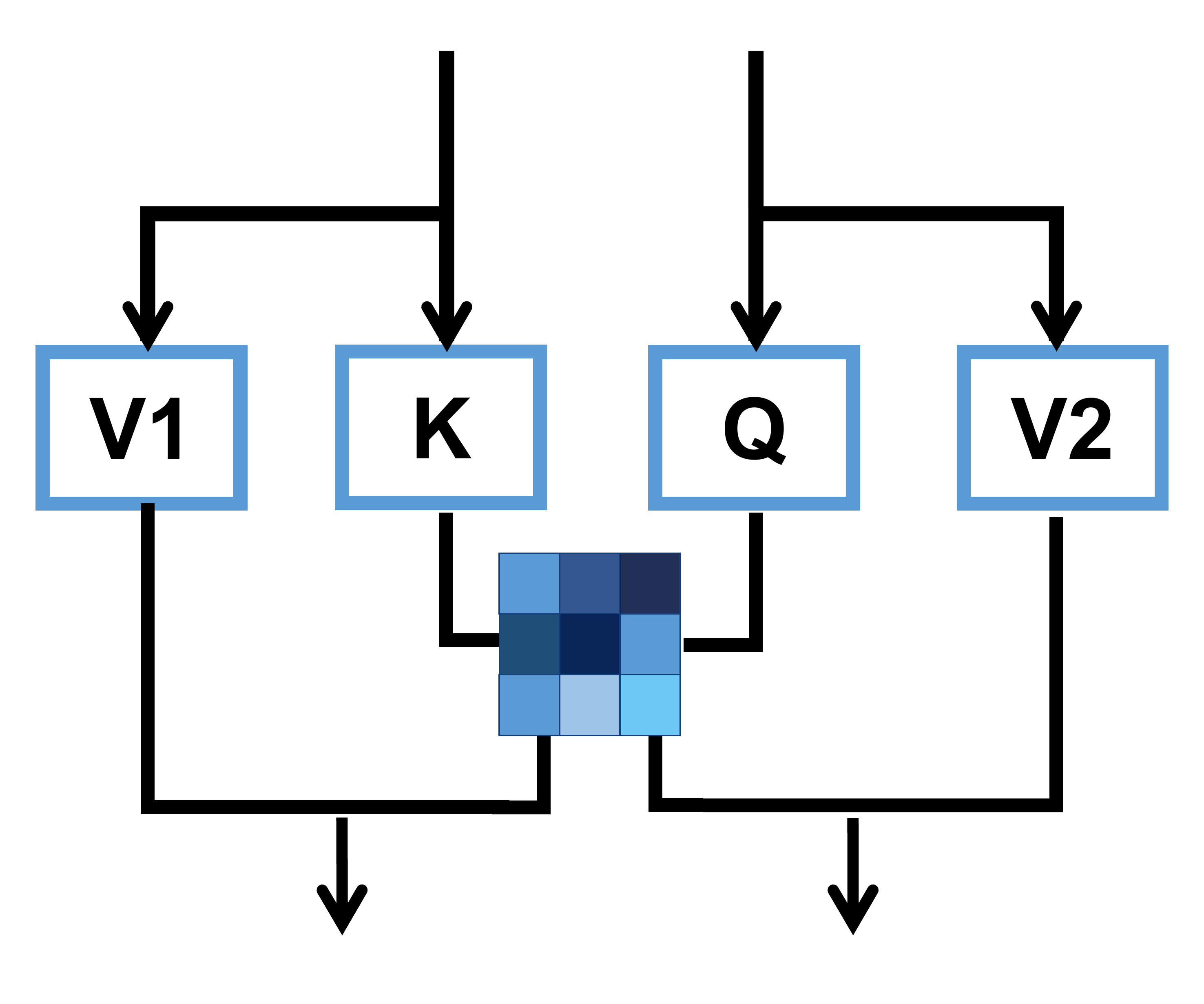}
    \label{fig:diff}
    }
    \caption{Illustration of three different attention units.}
    \label{fig:units}
    \vspace{-0.4cm}
\end{figure}

\subsection{Action-Background Interaction Module}


In this section, we describe the Action-Background Interaction (ABI) module in our BCNet.
The ABI module conducts action-background interaction on two temporal granularities: frame-level and clip-level. 
Inspired by the Transformer~\cite{vaswani2017attention}, ABI module models the interactions with two attention units, \emph{i.e.}, self-attention unit and difference-attention unit.
The details are described next.


\textbf{Frame-level Interaction.} We obtain the frame-wise feature sequence 
$F_o \in \mathbb{R}^{T\times C}$ via the feature encoder, where $T$ is the length of the feature sequence and $C$ is the feature dimension.
First, we use the \emph{Self-attention Unit} to learn the relationships between frames and enhance the feature representations.
The structure of the \emph{Self-attention Unit} is shown in Figure~\ref{fig:self}.
We utilize the Transformer-alike structure to devise the self-attention unit, which consists of two sub-layers: self-attention layer and feed-forward network (FFN).
Specifically, the input sequence $F_o$ is projected onto three learnable linear transformations to get queries $Q_{F_o}$, keys $K_{F_{o}}$, and values $V_{F_o}$.
Then the self-attention map can be calculated as follows:
\begin{equation}
    A(Q_{F_o},K_{F_o}) = Softmax \left( \frac {Q_{F_{o}}{K_{F_{o}}}^T}{\sqrt{C}} \right).
\end{equation}
We updates each component of the sequence by aggregating global information from the complete frame sequence by 
\begin{equation}
     F_{att}  = \varphi^g \left(  A(Q_{F_o},K_{F_o}) V_{F_{o}} \right),
\end{equation}
where $\varphi^g$ is a linear projection function.
Also, a residual connection around each of the two sub-layers and layer normalization is adopted to generate enhanced feature $F$, which can be written as:
\begin{equation}
\begin{split}
F^{'} = LayerNorm \left( F_{att} + F_o \right), \\
F = LayerNorm \left( F^{'} + FFN(F^{'}) \right). 
\end{split}
\end{equation}

Next, the enhanced features $F$ are fed into the \emph{Difference-attention Unit} to aggregate features based on the inconsistency between the action feature $F_a$ and the background feature $F_b$ under action and background supervision.
The structure of the \emph{Difference-attention Unit} is shown in Figure~\ref{fig:diff}.
The action feature $F_a$ is projected onto two learnable linear transformations to get queries $Q_{F_a}$ and values $V_{F_a}$.
Also, we transform the background features $F_b$ to queries $K_{F_b}$ and values $V_{F_b}$.
Then, we compute the difference map $A(Q_{F_{a}}, K_{F_{b}})$ as:
\begin{equation}
    A(Q_{F_{a}}, K_{F_{b}}) = Softmax \left( \frac {Q_{F_{a}}{K_{F_{b}}}^T}{\sqrt{C}} \right).
\end{equation}
In this way, $A_{i, j}$ represents the difference between the action frame $i$ and the background frame $j$. The smaller the value of $A_{i, j}$, the bigger the difference between the background features and action features.
Then we use the difference map to reweight $V_{F_a}$, $V_{F_b}$, and obtain enhanced feature $F_{a}^{'}$, $F_{b}^{'}$ respectively.

Finally, we append a prediction head which encode the $F_{a}^{'}$,$F_{b}^{'}$ with multi-layer perceptron (MLP) network and followed by a \emph{Sigmoid} layer to generate the action and background probability sequence.

\textbf{Clip-level Interaction.} We first use sliding window group to generate the action anchors and background anchors with different scales. 
Following BMN~\cite{lin2019bmn}, we construct weight term $w_{i,j} \in \mathbb{R}^{N \times T}$ via uniformly sampling $N$ points between the temporal region for each anchor. 
First, we conduct dot product in temporal dimension between $w_{i,j}$ and ${F_{a}^{'}}$  with the shape ${C \times N}$ to generate the action anchor.
Then, we get action anchor sequence $F_{a}^{c} \in \mathbb{R}^{L \times S}$ where $L$ is number of clip and $S = C \times N$. 
Similarly, we generate background anchors sequence $F_{b}^{c}$ in the same way. 

Next, the anchor sequences $F_{a}^{c}$ and $F_{b}^{c}$ are fed into the clip-level action-background interaction to generate action score and background score.
Specifically, we first utilize two independent \emph{Self-attention Units} to capture the relationships among action/background anchors, respectively.
The two self-attention units output updated anchor sequence ${F_{a}^{c}}^{'}$ and ${F_{b}^{c}}^{'}$.
Similar to the frame-level interaction, ${F_{a}^{c}}^{'}$ and ${F_{b}^{c}}^{'}$ are then fed into the \emph{Difference-attention Unit} to obtain difference map and reweighted anchor sequence $\widetilde{{F_{a}^{c}}^{'}}$ and  $\widetilde{{F_{b}^{c}}^{'}}$.
Note that the difference map $A_{i, j}$ represents the feature difference between $i$-th action anchor and $j$-th background anchor.

Finally, we add a clip-level predictor which encodes the $\widetilde{{F_{a}^{c}}^{'}}$ and $\widetilde{{F_{b}^{c}}^{'}}$  with multi-layer perceptron (MLP) and a $Sigmoid$ layer to predict action scores and background scores.


\subsection{Boundary Prediction Module}

Long-term temporal modeling is a critical factor in proposal boundary prediction. 
It is natural to use self-attention mechanism to model long dependencies. 
However, global modeling is easy to introduce global noise then leads to the over-smoothing.
To this end, we propose a \emph{Boundary Prediction} (BP) module which introduces original features to alleviate this phenomenon.
This module is built using the Transformer-alike structure which consist of multiple layers. Each layer contains a \emph{Self-attention Unit}, a \emph{Cross-attention Unit} and a feed-forward network.
Specifically, we first obtain the feature $F_i$ ($i$ represents the input features of layer $i$, if $i$ = 1, $F_i$ = $F_o$) and $F_o$.
Then, we feed them to the \emph{Self-attention Unit} and generate augmented global features $F_{i}^{g}$ and $F_{o}^{g}$. 
As shown in Fig.~\ref{fig:cross}, we use \emph{Cross-attention Unit} to generate the attention map $A(F_{i}^{g},F_{o}^{g})$ which represents the similarity between the aggregated feature $F_{i}^{g}$ and the original aggregated feature $F_{o}^{g}$, called the originality score.
To get $F_{i+1}$, we aggregate features which have high originality scores and discard features which have low originality scores.



The final output representation is then used as the global representation for the boundary prediction task. Specifically, we utilize a boundary predictor which encode the output representation with multi-layer perceptron (MLP) network and followed by a \emph{Sigmoid} layer to generate boundary probability sequence.

\subsection{Training}
\label{sec:train}
The overall objective of our framework is defined as:
\begin{equation}
    \mathcal{L} = \mathcal{L}_{1} + \mathcal{L}_{2},
\end{equation}
where $\mathcal{L}_1$ and $\mathcal{L}_2$ are the objective functions of the ABI module and the BP module respectively.

\textbf{Objective of BP module.}
The BP module generates the starting and ending probability sequence $P_s$, $P_e$.
Thus, the loss function consists of starting loss and ending loss:
\begin{equation}
    \mathcal{L}_{1} = \mathcal{L}_{bl}(P_s, G_s) + \mathcal{L}_{bl}(P_e, G_e),
\end{equation}
where $G_s$ and $G_e$ are the ground truth labels of boundary sequence, and $\mathcal{L}_{bl}$ is the binary logistic regression loss.

\textbf{Objective of ABI module.}
The ABI module generates frame-level and clip-level scores: $P_{a}^{f}$,$P_{b}^{f}$,$\widehat{P_{a}^{c}}$,$\widetilde{P_{a}^{c}}$ and $P_{b}^{c}$.
$P_{a}^{f}$ and $P_{b}^{f}$ are frame-level action and background classification scores.
$P_{b}^{c}$ is clip-level background classification scores.
Following BMN, $\widehat{P_{a}^{c}}$ is clip-level action classification scores and  $\widetilde{P_{a}^{c}}$ is regression action scores.
The loss function $\mathcal{L}_{2}$ consists of frame-level loss and clip-level loss:
\begin{equation}
    \mathcal{L}_{2} = \mathcal{L}_{frame} + \mathcal{L}_{clip}.
\end{equation}
The frame-level loss is 
\begin{equation}
    \mathcal{L}_{frame} = \mathcal{L}_{c}(P_{a}^{f},G_a^f) + \mathcal{L}_{c}(P_{b}^{f},G_b^f),
\end{equation}
where $G_a^f$ and $G_b^f$  are the ground truth labels of action and background probability at frame-level.
The clip-level loss is formulated as follows:
\begin{equation}
    \mathcal{L}_{clip} = \mathcal{L}_c(\widehat{P_{a}^{c}},G_a^c) + \mathcal{L}_r(\widetilde{P_{a}^{c}},G_a^c) +
    \mathcal{L}_c(P_{b}^{c},G_b^c),
\end{equation}
where $G_a^c$ and $G_b^c$ are the ground truth labels of action and background scores at clip-level. 
$\mathcal{L}_{c}$ denotes the binary logistic regression loss function and $ L_r$ is a smooth $L_1$ loss. 

\subsection{Inference}
\label{sec:infer}
As mentioned above, the BP module generates boundary probability and the ABI module generates the action and background scores.
Then we take the boundary probability, action scores and background scores into the Post-processing module.
Firstly, we construct a proposals set $\psi^c_p$ based on boundary probabilities.
Second, the proposal is refined by a corresponding pre-set anchor.
The proposal $\varphi = [t_s',t_e'] \in {\psi}^c_p$ is taken as an example, 
we compute the temporal Intersection over Union (tIoU) between proposal $\varphi$ and anchors, then select a matching anchor $p_m = [t^m_s, t^m_e]$ to refine proposals.
we refine the proposal as:
\begin{equation}
   [t_s,t_e]=
    \begin{cases}
         [\frac{ t_s' + t^m_s}{2} , \frac{ t_e' + t^m_e}{2}], & if ~ {\widehat{p_{m}^{a}} > {\alpha}_1} ~ and ~ \widetilde{{p_{m}^{a}}} > {\alpha}_2 \\
         [t_s' ,  t_e'], &others
    \end{cases},
\end{equation}
where $\widehat{p_{m}^a}$ is the anchor action classification score, $\widetilde{p_{m}^a}$ is the action regression score, ${\alpha}_1$ and ${\alpha}_2$ are the adjustment thresholds.
Finally, we get a proposal set $\psi_p = \{{\phi}_n = (t_s,t_e,p_{t_s'}^{s},p_{t_e'}^e,\widehat{p_{m}^a},\widetilde{p_{m}^a},{p_{m}^b})\}_{n=1}^N$, where $p_{t_s'}^{s},p_{t_e'}^{e}$ are the starting and ending probabilities and ${p_{m}^b}$ is anchor background score.

Following the previous practices, we also perform score fusion and redundant proposal suppression to further obtain final results.
Specifically, in order to make full use of various predicted scores for each proposal $\varphi_n$, we fuse its boundary probabilities and action-background scores of matching anchor by multiplication.
The confidence score $p^f$ can be defined as :
\begin{equation}
    p^f = p_{t_s'}^s \cdot p_{t_e'}^e \cdot \widetilde{p_{m}^a} \cdot  \widehat{p_{m}^a} \cdot (1 - {p_{m}^b}).
\end{equation}
Hence, the final proposal set as 
\begin{equation}
    \psi = {\{\varphi_n = (t_s,t_e,p^f)\}}_{n=1}^N.
\end{equation}
Moreover, we also use the Soft-NMS algorithm for post-processing to remove the proposals which highly overlap with each other.

\section{Experiments}
\subsection{Datasets and Evaluation Metrics}
\textbf{ActivityNet-v1.3} ~\cite{activitynet} is a large-scale video dataset for action recognition and temporal action detection tasks.
It contains 10K training, 5k validation, and 5k testing videos with 200 action categories, and the ratio of training, validation and testing sets is 2:1:1.
\textbf{THUMOS14}~\cite{thumos} contains 200 validation untrimmed videos and 213 test untrimmed videos, including 200 action categories.
This dataset is challenging due to the large variations in the frequency and duration of action instances across videos.

\noindent\textbf{Evaluation Metrics. } 
Temporal action proposal generation aims to produce high-quality proposals with high tIoU, which have a high recall rate. 
To evaluate quality of proposals, Average Recall (\textbf{AR}) is the average recall rate under specified tIoU thresholds.
Following the standard protocol, we use thresholds set [0.5:0.05:0.95] on ActivityNetv1.3 and [0.5:0.05:1.0] on THUMOS14.
To evaluate the performance of temporal action detection task, mean Average Precision (\textbf{mAP}) under multiple tIoU is the widely-used evaluation metric.
On ActivityNet-v1.3, the tIoU thresholds are set to \{0.5, 0.75, 0.95\}, and we also test the average mAP of tIoU thresholds between 0.5 and 0.95 with step of 0.05.
On THUMOS14, these tIoU thresholds are set to \{0.3, 0.4, 0.5, 0.6, 0.7 \}.

\begin{table}[ht]
\centering
\scalebox{0.95}{
\begin{tabular}{cccccc}
\shline
 Method &@50 &@100 &@200 &@500 &@1000 \\ \hline
 TAG~&18.6 & 29.0 & 39.6 & - & - \\
 CTAP~ &32.5 & 42.6 & 52.0 & - & - \\
 BSN~& 37.5 & 46.1 & 53.2 & 61.4 & 65.1 \\
 MGG~&39.9 & 47.8 & 54.7 &61.4 & 64.6 \\
 BMN~&39.4 & 47.7 & 54.8 & 62.2 & 65.5 \\
 BSN++ ~&42.4 & 49.8 & 57.6 & 65.2 & 66.8 \\
 TCANet~&42.1 & 50.5 & 57.1 & 63.61 & 66.9 \\
 RTD-Net~& 41.1 & 49.0 &56.1 & 62.9 & - \\ \hline
\textbf{Ours} & \textbf{45.5} & \textbf{53.6} & \textbf{60.0} & \textbf{67.0} & \textbf{69.8} \\
\shline
\end{tabular}
}
\caption{Performance comparison with state-of-the-art proposal generation methods on test set of THUMOS14 in terms of AR@AN.}
\label{tab:thumos_ar_an}
\end{table}

\begin{table}[th]
\centering
\scalebox{0.86}{
\begin{tabular}{ccccccc}
\shline
Method& Feature &0.3 &0.4 &0.5 &0.6&0.7 \\ \hline
SST~&TSN& - & - &23.0 &- &- \\
TURN~& TSN&44.1 & 34.9 & 25.6 & - &- \\
SSN~& TSN &51.9 & 41.0 &29.8 &- &- \\
BSN~& TSN & 53.5 & 45.0 &36.9 &28.4 &20.0 \\
MGG~& TSN &53.9 & 46.8 &37.4 &29.5 &21.3 \\
DBG~& TSN &57.8 & 49.4 &39.8 &30.2 &21.7 \\
BMN~& TSN &56.0 & 47.4 &38.8 &29.7 &20.5 \\
G-TAD~& TSN &54.5 & 47.6 &40.2 &30.8 &23.4 \\
BSN++~& TSN &59.9 & 49.5 &41.3 &31.9 &22.8 \\ 
TCANet~& TSN & 60.6 & 53.2 &44.6 &36.8 &26.7 \\\hline
\textbf{Ours} & TSN & 66.5& 60.0 & 51.6 & 41.0 & 29.2\\
\hline\hline
BU-TAL~& I3D &53.2 & 48.5 &42.8 &33.8 &20.8 \\
P-GCN ~& I3D &63.6& 57.8& 49.1 & - & -\\
AFSD ~ & I3D &67.3& 62.4 & 55.5 & 43.7 & 31.1 \\
RTD-Net~& I3D &53.9 & 48.9 & 42.0 & 33.9 & 23.4 \\
G-TAD+P-GCN ~ &I3D& 66.4 & 60.4 & 51.6 & 37.6 &22.9\\
RTD+P-GCN ~ &I3D& 68.3& 62.3 & 51.9 & 38.8 & 23.7\\
MUSES ~ & I3D & 68.9& 64.0 & 56.9 & 46.3 & 31.0\\ \hline
\textbf{Ours}+P-GCN~ &I3D& 69.8 & 62.9& 52.0 & 39.8 & 24.0\\
\textbf{Ours}+MUSES ~ &I3D& \textbf{71.5}&\textbf{ 67.0 }& \textbf{60.0} & \textbf{48.9} & \textbf{33.0}\\


\shline
\end{tabular}
}
\caption{Performance comparison with state-of-the-art action detection methods on test set of THUMOS14, in terms of mAP (\%) at different tIoU thresholds.}
\label{tab:thumos_map}
\end{table}

\begin{table}[th]
\centering
\scalebox{0.95}{
\begin{tabular}{cccc}
\shline
Method & AR@1 (val) & AR@100 (val) &AUC (val) \\ \hline
CTAP~ & - & 73.2  & 65.7 \\
BSN~ & 32.2 & 74.2 & 66.2 \\
MGG~ & - & 75.5 & 66.4 \\
BMN~& - & 75.0 & 67.0 \\
BSN++~& 34.3 & 76.5 & 68.3 \\
TCANet ~& 34.6 & 76.1 & 68.1 \\
RTD-Net~& 32.8 & 73.1&65.7\\ \hline
\textbf{Ours} & \textbf{35.2} & \textbf{76.6} & \textbf{68.7} \\
\shline
\end{tabular}
}
\caption{Performance comparison with state-of-the-art proposal generation methods on validation set of ActivityNet-1.3 in terms of AUC and AR@AN.}
\label{tab:anet_ar_an}

\end{table}

\begin{table}
\centering
\scalebox{0.95}{
\begin{tabular}{ccccc}
\shline

Method  &0.5 &0.75 &0.95 &Average \\ \hline
Singh et al.& 34.5 &- &- &- \\
SCC~& 40.0 & 17.9 & 4.7 & 21.7 \\
CDC~& 45.3 & 26.0 & 0.20 &23.8 \\
R-C3D~& 26.8 &- &- &- \\
BSN~& 46.5& 30.0 & 8.0 & 30.0 \\
BMN~& 50.1 & 34.8 & 8.3 & 33.9 \\
GTAD~& 50.4 & 34.6 & 9.0 & 35.1 \\
BSN++~& 51.3 &35.7& 8.3 & 34.9 \\
TCANet \emph{w/} BSN~& 51.9 & 34.9 & 7.5 & 34.4 \\
RTD-Net~&46.4 & 30.5 & 8.6 & 30.5 \\ \hline
\textbf{Ours} & \textbf{53.2} & \textbf{36.2 }& \textbf{10.6} & \textbf{35.5}\\

\shline
\end{tabular}
}

\caption{Performance comparison with state-of-the-art action detection methods on validation set of ActivityNet-1.3, in terms of mAP (\%) at different tIoU thresholds and the average mAP.}
\label{tab:anet_map}
\end{table}

\subsection{Implementation Details}
\textbf{Feature Encoding.} Following previous works~\cite{lin2019bmn,xu2020g}, we adopt the TSN~\cite{tsn} and I3D \cite{two-stream} for feature encoding.
For THUMOS14, the interval $\sigma$ is set to 8 and 5 for I3D and TSN respectively. We crop each video feature sequence with overlapped windows of size $T = 256$ and stride 128.
As for ActivityNet-1.3, the sampling frame stride is 16, and each video feature sequence is rescaled to $T = 100$ snippets using linear interpolation.
\\
\textbf{Training and Inference.} 
The number of layers in Boundary Prediction module is 12.
Due to the limit of computation resource, we apply 1D Conv for dimension reduction, then take the features as the input to the Boundary Prediction module and Action-Background Interaction module.
For each anchor, we use sampling points $N = 32$.
For post-processing module, we set adjustment thresholds ${\alpha}_1 = 0.9$  and ${\alpha}_2 =0.8 $.
We train our model from scratch using the Adam optimizer and the learning rate is set to $10^{-4}$ and decayed by a factor of 0.1 after every 10 epoch.

\subsection{Comparison with State-of-the-arts}
Here we compare our BCNet with the existing state-of-the-art methods on ActivityNet-v1.3 and THUMOS14.
For fair comparisons, we adopt the same two-stream features used by previous methods in our experiments.

\textbf{Results on THUMOS14.} 
BCNet is compared with state-of-the-art methods in  Table~\ref{tab:thumos_ar_an} and Table~\ref{tab:thumos_map}, where our method improves the performance significantly for both temporal action proposal generation and action detection.
For the temporal action proposal generation task, results are shown in Table~\ref{tab:thumos_ar_an}, which demonstrate that BCNet outperforms state-of-the-art methods in terms of AR@AN with AN varying from 50 to 1000.
\begin{table*}[t]
\centering
\begin{tabular}{ccccccc}
\shline
Method & w/ BC  & 0.3 & 0.4 & 0.5 & 0.6 & 0.7 \\ \hline
BMN* & - & 59.5 & 54.3 & 45.1 & 35.3 &  24.8 \\
BMN* & \Checkmark & 62.5 (\textcolor{red}{$\uparrow$ 3.0}) & 56.3  (\textcolor{red}{$\uparrow$ 2.0}) & 47.6 (\textcolor{red}{$\uparrow$ 2.5}) & 37.2 (\textcolor{red}{$\uparrow$ 1.9}) & 26.3 (\textcolor{red}{$\uparrow$ 1.5}) \\ \hline
GTAD* & -  & 58.4 & 52.1 & 43.5 & 33.3 & 23.2 \\
GTAD* & \Checkmark & 60.5 (\textcolor{red}{$\uparrow$ 2.1}) & 54.1 (\textcolor{red}{$\uparrow$ 2.0}) & 45.7 (\textcolor{red}{$\uparrow$ 2.2}) & 35.1 (\textcolor{red}{$\uparrow$ 1.8}) & 24.5 (\textcolor{red}{$\uparrow$ 1.3}) \\ \hline
Ours & - & 63.2 & 58.7 & 51.2 & 39.9 &  28.3\\
Ours & \Checkmark & 66.5 (\textcolor{red}{$\uparrow$ 3.3}) & 60.0 (\textcolor{red}{$\uparrow$ 1.3}) & 51.6 (\textcolor{red}{$\uparrow$ 0.4}) & 41.0 (\textcolor{red}{$\uparrow$ 1.1}) & 29.2 (\textcolor{red}{$\uparrow$ 0.9}) \\
\shline
\end{tabular}

\caption{The effectiveness of the Background Constraint (BC). * indicates our implementation with the publicly available code.}
\label{tab:ablation_background}
\end{table*}

For the temporal action detection task, the proposed BCNet also achieves superior results, as shown in Table~\ref{tab:thumos_map}. 
The performance of our method exceeds state-of-the-art proposal generation methods by a big margin at different tIoU thresholds.
Specially, BCNet based on TSN feature reaches an mAP of 51.6 $\%$ at IoU 0.5.
Besides, the performance of BCNet can be further boosted when
it is combined with proposal post-processing methods: P-GCN \cite{zeng2019graph} and MUSES~\cite{liu2021multi}.
Now BCNet reaches \textbf{60.0\%} at IoU 0.5, outperforming all the other methods. 
This signifies the advantage of BCNet proposals regardless of post-processing.


\textbf{Results on ActivityNet-v1.3.} 
In Table~\ref{tab:anet_ar_an} and Table~\ref{tab:anet_map}, we compare the proposed BCNet with other methods on ActivityNet-v1.3.
For the temporal action proposal generation task, as shown in Table~\ref{tab:anet_ar_an}, the performance of BCNet again outperforms state-of-the-art proposal generation methods in terms of AR@AN with AN varying from 1 to 100 and AUC.
Especially when AN equals 1, we achieve 35.2\% regarding the AR metric, which indicates that top-1 proposal has high quality.
For the temporal action detection task, as summarized in Table~\ref{tab:anet_map},
our method achieves notable improvements on mAP over other proposal generation methods such as BMN~\cite{lin2019bmn} and G-TAD~\cite{xu2020g} at all tIoU thresholds.
When tIoU is 0.95, the mAP we obtain is 10.6\%, indicating that the confidence of the generated proposals are more reliable.

\subsection{Ablation Study}

In this section, we conduct ablation studies on THUMOS14 to verify the effectiveness of each component in BCNet.

\textbf{Multi-level ABI module.}
We perform ablation studies to verify the effectiveness of multi-level interaction in ABI module.
Frame-level interaction is designed to generate features of action and background.
Here, the ablation experiment demonstrates the necessity of frame-level interaction as shown in Table \ref{tab:ablation_multi}.
Compared with single-level ABI module that only has a clip-level interaction, multi-level ABI module is improved by 3.6\% at tIou 0.3.


\begin{table}[th]
\centering
\scalebox{0.95}{
\begin{tabular}{cccccccc}
\shline
Frame & Clip  &0.3 &0.4 &0.5 &0.6&0.7 \\ \hline
- & \Checkmark & 62.9& 57.9 & 49.6 & 40.4 & 28.6 \\

\Checkmark & \Checkmark & 66.5 & 60.0 & 51.6 & 41.0& 29.2 \\
\shline
\end{tabular}
}
\caption{The effect of ABCNet in frame-level and clip-level.}
\label{tab:ablation_multi}
\vspace{-0.2cm}
\end{table}

\textbf{The effectiveness of Background Constraint.}
We perform ablation studies to verify the effectiveness of the background constraint idea.
To validate the generalizability of our proposed background constraint idea, we add it to the BMN, GTAD.
The experimental results are shown in Table ~\ref{tab:ablation_background}, which reveals that background constraint can also significantly improve the performance of existed methods.


\textbf{Architecture of ABI module.}
We perform ablation studies to verify the effectiveness of the architecture of ABI module.
To generate reliable confidence of proposal, ABI module is designed by exploiting rich information of action and background. 
Our proposed ABI module consists of two key units: self-attention unit and difference-attention unit.
Results are shown in Table \ref{tab:ablation_ABI}.
The self-attention unit can improve the performance by a large margin (almost 4.5$\%$) at tIoU 0.5.
Difference-attention unit also brings significant improvement at tIoU 0.3, as the inconsistency of action and background is captured between action and background.
\begin{table}[th]
\centering
\scalebox{1}{
\begin{tabular}{ccccccc}
\shline
Self.&Diff. &0.3 &0.4 &0.5 &0.6&0.7 \\ \hline
- & - & 59.3 & 55.8 & 46.7 & 36.6 &  25.2\\
\Checkmark & - &63.2 & 58.7 & 51.2 & 39.9 &  28.3\\
\Checkmark & \Checkmark &66.5 & 60.0& 51.6 & 41.0 & 29.2 \\
\shline
\end{tabular}
}
\caption{The effectiveness of ABI module.}
\label{tab:ablation_ABI}
\end{table}

\subsection{Analysis on runtime.}
To verify the efficiency of our BCNet, we report the latency of our method on THUMOS14.
For the fair comparisons with other models, we measure the latency under the same environment (a single NVIDIA 2080Ti GPU). 
We use a batch size of 1 to measure the latency on the full testing set and report the average time.
As shown in Table~\ref{ablation_infer}, our BCNet achieve the best mAP with smallest latency (141ms \emph{v.s.} 298ms, 330ms). 
The main reason is that our model generates fewer proposals than these methods, which helps our model run faster.
\begin{table}[th]
\centering
\scalebox{1}{
\begin{tabular}{cccccccc}
\shline
Method  &0.3 &0.4 &0.5 &0.6&0.7 & Latency \\ \hline
BMN & 56.0 & 47.4 & 38.8 & 29.7 &  20.5 & 330ms\\
GTAD & 54.5& 47.6 & 40.2 & 30.8 &  23.4 & 298ms\\
Ours  & 66.5 & 60.0 & 51.6 & 41.0 &  29.2& 141ms\\

\shline
\end{tabular}
}

\caption{Quantitatively analysis on latency. The smaller latency represent higher efficiency.}
\label{ablation_infer}
\end{table}

\section{Conclusion}
In this paper, we introduce a Background Constraint concept, which can be integrated easily with existing TAPG method.
Based on this concept, we propose a Background Constraint Network, which consists of multiple attention units \emph{i.e.}, self-attention unit, cross-attention unit, and difference-attention unit, and generates high-quality proposals by exploiting inconsistency between action and background.
Extensive experiments show that our model achieves new state-of-the-art performance in temporal action proposal generation and action detection on THUMOS14 and ActivityNet1.3 datasets.

\section*{Acknowledgments}
The work was funded by National Natural Science Foundation of China (U1711265) and Heilongjiang Province Science Foundation (2020ZX14A02).

\bibstyle{aaai22}
\bibliography{aaai22}
\end{document}